\documentclass[conference]{IEEEtran}
\IEEEoverridecommandlockouts
\usepackage{cite}
\usepackage{amsmath,amssymb,amsfonts}
\usepackage{algorithmic}
\usepackage{graphicx}
\usepackage{textcomp}
\usepackage{xcolor}
\usepackage{todonotes}
\usepackage{url}
\def\BibTeX{{\rm B\kern-.05em{\sc i\kern-.025em b}\kern-.08em
    T\kern-.1667em\lower.7ex\hbox{E}\kern-.125emX}}
\begin{document}

\title{An Aircraft Upset Recovery System with Reinforcement Learning\\
}

\makeatletter
\newcommand{\linebreakand}{%
  \end{@IEEEauthorhalign}
  \hfill\mbox{}\par
  \mbox{}\hfill\begin{@IEEEauthorhalign}
}
\makeatother

\author{
\IEEEauthorblockN{Mahir Demir}
\IEEEauthorblockA{
\textit{Turkish Aerospace}\\
Istanbul, Turkey \\
mahir.demir@tai.com.tr}
\and
\IEEEauthorblockN{Atahan Cilan}
\IEEEauthorblockA{
\textit{Turkish Aerospace}\\
Istanbul, Turkey \\
atahan.cilan@tai.com.tr}
\and
\IEEEauthorblockN{Seyyid Osman Sevgili}
\IEEEauthorblockA{
\textit{Turkish Aerospace}\\
Istanbul, Turkey \\
seyyidosman.sevgili@tai.com.tr}
\and
\IEEEauthorblockN{Ozgun Can Yurutken}
\IEEEauthorblockA{
\textit{Turkish Aerospace}\\
Istanbul, Turkey \\
ozguncan.yurutken@tai.com.tr}
\linebreakand
\IEEEauthorblockN{Umit Can Bekar}
\IEEEauthorblockA{
\textit{Turkish Aerospace}\\
Istanbul, Turkey \\
can.bekar@tai.com.tr}
}

\maketitle

\begin{abstract}
This article explores the progress made in the creation of a pilot activated recovery system (PARS) for advanced jet trainers that utilizes artificial intelligence (AI) in an effort to enhance operational efficiency. The PARS model employs an advanced reinforcement learning (RL) architecture, incorporating a cutting-edge soft-actor critic (SAC) model and hyper-parameter optimization methods. Negative-g punishments and other handcrafted features remarked upon by control engineers and domain experts regarding PARS are also taken into account by the system. When evaluated by them, the AI model's behavior is deemed more desirable than that of conventional control methods. 
\end{abstract}


\section{Introduction}
This research shows that artificial intelligence (AI) is capable of efficiently resolving challenges associated with upset recovery problem, and gives the historical PARS solution a new look. The upset recovery problem is generally associated with stall and spin conditions, and is the most common reason for the aircraft accidents \cite{ReportBoeing}

As indicated previously, the SAC algorithm is employed to train the PARS. The algorithm is provided with the properties of advanced jet trainers and the command errors, which are to be integrated into the action space so that oscillations associated with those actions can be eliminated. The RL agent possesses the capability to compute the elevator and aileron degrees.

The principal challenge encountered in our investigation pertains to the formulation and implementation of a suitable incentive function for PARS that fulfills the necessary criteria. A multitude of reward function design schemes were examined during the development phase, starting with gamma($\gamma$) angle and a zero-degree roll($\phi$) as the exclusive targets. Subsequently, to resolve the oscillation concern that emerged, we formulated a penalty for derivative command errors.

Furthermore, we have expanded our reinforcement learning model by incorporating the sequential reward technique, whereby the significance of the rewards or errors associated with a particular property increases as the associated property approaches its target. We have experimented with numerous property associations using this sequential scheme in an effort to identify the optimal reward design. Furthermore, g errors have been incorporated to ensure that the pilot is not adversely affected by an excessive amount of negative g force.

Our final model, which has been evaluated by domain experts and simulation data, demonstrates that the implementation of AI significantly improves the performance of PARS while reducing the amount of labor-intensive control engineering required. The significance of AI in the field of avionics is validated by these results, which also demonstrate that these types of systems have the potential to surpass traditional control mechanisms. Modern RL pipelines were integrated into our work, along with a reward and punishment system applicable to other similar flight control tasks.

In summary, our research enhances the traditional control-based PARS for an advanced jet trainer by incorporating modern RL methodologies. In order to address this, a contemporary reinterpretation of the RL baseline and reward/punishment systems tailored to the aerospace engineering domain are devised. The outcomes are assessed in collaboration with the domain experts of said aircraft and will facilitate subsequent research aimed at furthering the field of air travel.

In our work, we have 4 main contributions:

\begin{itemize}
  \item We have used a high-fidelity aircraft flight dynamics model in the simulation environment to develop and test our algorithm, and validated our resulting PARS model with domain experts in the simulation environment. That way, we have trained our algorithm for possible upset conditions (e.g. stall, spin), while traversing a discretized state space randomly.
  \item We have developed the first algorithm that is fully independent from traditional control algorithms with using RL without any external constraints, giving the actor agent a free space to find otherwise not possible through optimal recovery maneuvers.
  \item We have modified and used state of the art SAC algorithm for an avionic manner for the first time.
  \item Our model was tested and evaluated with both domain experts in the simulator and control engineers, giving it credibility for being an efficient PARS algorithm implementation.
\end{itemize}

The rest of the paper is organized as follows:

In the second section, a short history of previous works are given. In the third section, an introduction to RL was provided. In the fourth section, the methodology is explained. The fifth section gives the simulation results. The sixth section concludes the paper. 

\section{Previous Works}

PARS is an important feature of a military aircraft flight control system. To implement this function, there are a lot of possible methods to use, however, due to safety critical requirements of avionic systems, actual avionics is being implemented via classical control mechanisms.

One such example is given in \cite{combs1992pilot}, where the authors developed control laws to handle dangerous flight conditions. To provide more refined recovery maneuvers, the authors first divided the parameter space into 7 sectors, which are defined by pitch and roll attitudes and, pitch and roll responses. Then, for each sector, necessary optimized recovery maneuvers were designed. The main constraint for this approach is the hand-crafted process of classical control design can not find the optimal solution. Though having the same shortcomings, a more robust classical control method is depicted in \cite{crespo2012analysis}, where authors had used linear controllers to handle the problem.

With the advancement of scholastic and learning-based methods, a variety of papers that based on such methods were introduced for PARS problem. 

In \cite{sweriduk1999design}, a genetic algorithm based PARS was presented. Authors had used 6-DoF aircraft model in a simulation environment. The genetic algorithm returns a control logic for PARS, which is evaluated with attitude, flight path angle, and acceleration command. The outputs of control logic are further constrained by some conditional statements. The long convergence time of genetic algorithms caused the research to not cover some possible flight envelopes like spin conditions. Compared with the classical methods, the genetic algorithm can find non-intuitive methods that are optimal, but still is not completely free due to hard constraints. 

A fuzzy logic approach is used in \cite{youssef1995fuzzy}. Based on \cite{combs1992pilot}, the classical control laws are used in the fuzzy manner. The results had shown that the performance was better and development costs were reduced. As this approach is based on \cite{combs1992pilot}, it still can not use freedom of learning-based models. 

The inclusion of advanced physics and optimization techniques had provided some different methodologies in PARS. As an example of that, in \cite{paranjape2018optimization} authors have incorporated numerical optimization methods into a physics-based proxy function. The main objective of this proxy function is to minimize energy costs while recovering. They have used a single-point mass model akin to \cite{hospodavrflight}. This is a novel addition to classical control methods while sharing the same constraints. 

\cite{kim2017reinforcement} shows an RL approach for PARS problem in the UAVs and for spin conditions. The authors first used bifurcation analysis to get stable initial spin points. Then, they have two different RL models for angular rate recovery and altitude recovery. They have used the Q-learning methodology RL model in conjunction with the classical PID controller. Their results provided better timing than other methods. 

\cite{tomar2021automated} worked on two variants of stall conditions, in 105 different test cases. They have used two models in a sequential way, the first being DDPG and second being classical regression. They have successfully recovered stall in average between 70 to 300 seconds in different altitudes, with providing a level flight termination.

\cite{wang2024aircraft} provides a pilot assistance system(PAS) for a upset condition recovery. They have enlarged the previous researchers' envelope for the recovery and used PPO to bring the aircraft back to level flight. 

\cite{cao2021two} developed two-stage RL strategy using TD3. First, they have developed the pre-train stage for general upset recovery, then they fine-tuned it for the specific aircraft, Cesna in this case. They have provided a comparison with PID and shown successful recovery.

In \cite{lang2022deep} NASA Generic Transport Model aircraft was used to provide a upset recovery solution. Using PPO, the authors provide an RL algorithm that brings the airplane to the safe states. They have done around 100000 tests and RL provides a suitable solution.

An offline-online RL strategy is used in \cite{jiang2023safely}. The authors first used behavioral cloning to teach the model how human pilots recover stall positions, then used online RL to fine-tune the model.

To handle missing parts of the previous works, our work incorporates the whole state space of the airplaneand provides state-of-the-art RL algorithms with no hard constraints. The important -g condition for termination is handled. The results were compared with classical controllers and the increased performance was proven with -g safety for pilot.

\section{Background}

\subsection{Upset Conditions}

In \cite{icao} the upset condition is defined as the unintentional deviation from the normal flight envelope, in which general value of  more than 25 degree nose up or more than 10 degree nose down is accepted as the starting condition, along with more than 45 degree bank angle.

Altrough there are a lot of subsections of upset conditions, they are collected in two in the literature. 

First one is called the stall, where the angle of attack is increased above the limits of normal conditions, as such there is no enough lift generated to perform a stable flight \cite{OMRAN2011353}.

Second one is called spin, where two wings are experience asymmetrical amount of lift, resulting in a rotating descent in the shape of a helicoid \cite{malik2020review}. 

We have mentioned that, in [1], the parameter space was divided into 7 subsections. The figure \ref{upset}, taken from that paper also shows one definition of upset conditions according to [1]

\begin{figure}[htbp]
\centerline{\includegraphics[width=0.5\textwidth]{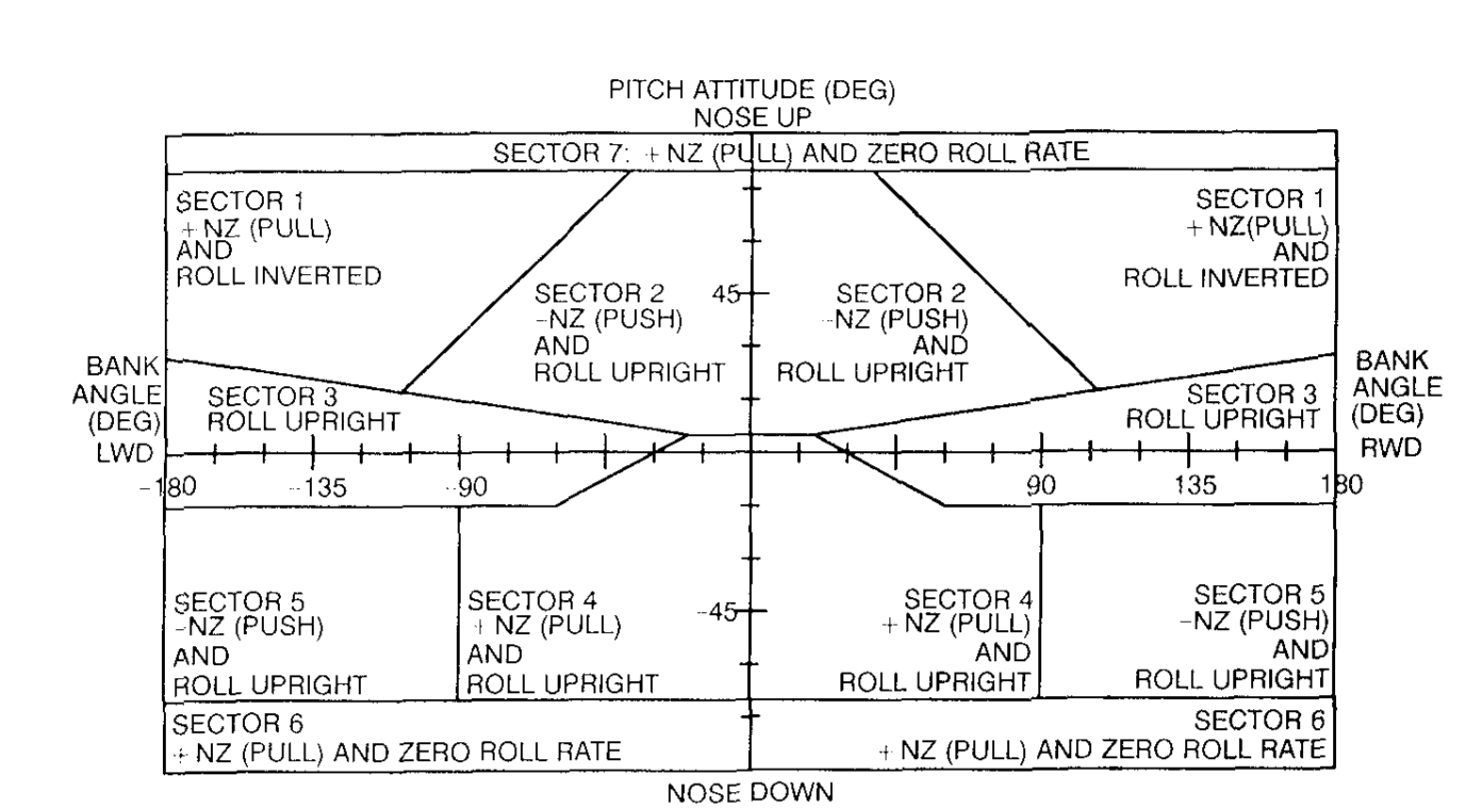}}
\caption{One description of upset condition sections, taken from [1]}
\label{upset}
\end{figure}

Given the formal definitions, due to end goal being to satisfy the pilot-activated recovery system, it is ultimately the pilot's objective sense to determine upset condition. Thus, in our objective, our objective function is to satisfy level flight, thus any condition that does not reflect that is the upset condition for us.

\subsection{Reinforcement Learning}

As we have mentioned previously, we are using the RL paradigm to learn aerobatic maneuvers for a desired aircraft. Thus, this section aims to provide a short overview of RL. 

Reinforcement learning is one of the three main learning methods of machine learning, the others being supervised learning and unsupervised learning. Reinforcement learning is explained as learning what to do -how to match the current state to the next action- so that that action maximizes a specific numerical reward. One more important difference is the actions to take are not given to the agent, like most of the other machine learning methods, thus it must learn them via trial and error. \cite{andrew2018reinforcement}. 

In RL, the agent, which is a mapping between observation and action space (generally referred as policy), observes a state $s_{t}$ from its environment at time step t. Then the agent decides on an action $a_{t}$. The action acts upon the environment and the total system shifts into the state $s_{t+1}$. The action will cause the environment to generate reward $R_{t+1}$\cite{arulkumaran2017brief}, which is the main feedback used for learning in RL. Generated data (state, action, reward pairs) is used to improve the current policy.

Figure \ref{fig} depicts an simple working principle of RL: 
\begin{figure}[htbp]
\centerline{\includegraphics[width=0.5\textwidth]{rl-algo.png}}
\caption{The general working principle of RL. States and rewards of previous step is inputted to agent, which generates an action acording to learned model. The action is given to environment, generating new state and reward.}
\label{fig}
\end{figure}

\section{Methodology}

This part discusses the PARS model development process with RL. Each part is build on top of previous part by improving encountered problems and changing the objective function (i.e reward function) to properly represent PARS mechanics. 

\subsection{Model and Environment Setup}

In our infrastructure, reward function includes several components and each component contributes to overall reward with proportional to a weight factor for that component. By introducing new reward components, changing the weight factor of each component or changing the normalization range (i.e scale factor), we can obtain various reward functions, namely an objective function. In the next section, we described the constructed reward function for each iteration and presented the results of this reward function.

Independent of the iteration, the logic of the reward function is the same. Depending on the error type, we are using 2 different reward scheme. Asymptotic and linear. The $\phi$ and $\gamma$  errors described in the next section are asymptotic, whereas the command errors are linear. 

For the asymptotic errors, the total reward is calculated via the following set of equations:

\[
e_\text{abs} = | x_{\text{target}} - x|,\text{ } 
\]
The first equation denotes the absolute error. This is same for both asymptotic and linear error components. It is direct difference between predetermined target value of the state and current value of the state. 

\[
e_{\text{scaled}} = \frac{e_\text{abs}}{scale}
\]
To satisfy asymptotic property, the scale factor contributes significant influence. The asymptotic property is used when it is needed for the system returning big error while having a great difference to the target, and small error if there is not so much difference. This aspect smooths the learning of RL model in some states, such as  $\phi$ and $\gamma$. 

However, making the curve asymptotic brings a careful design on scaling factor since a random factor may miss the 0-1 normalization range in the reward. The figure \ref{scale} shows this in roll example:

\begin{figure}[htbp]
\centerline{\includegraphics[width=0.50\textwidth]{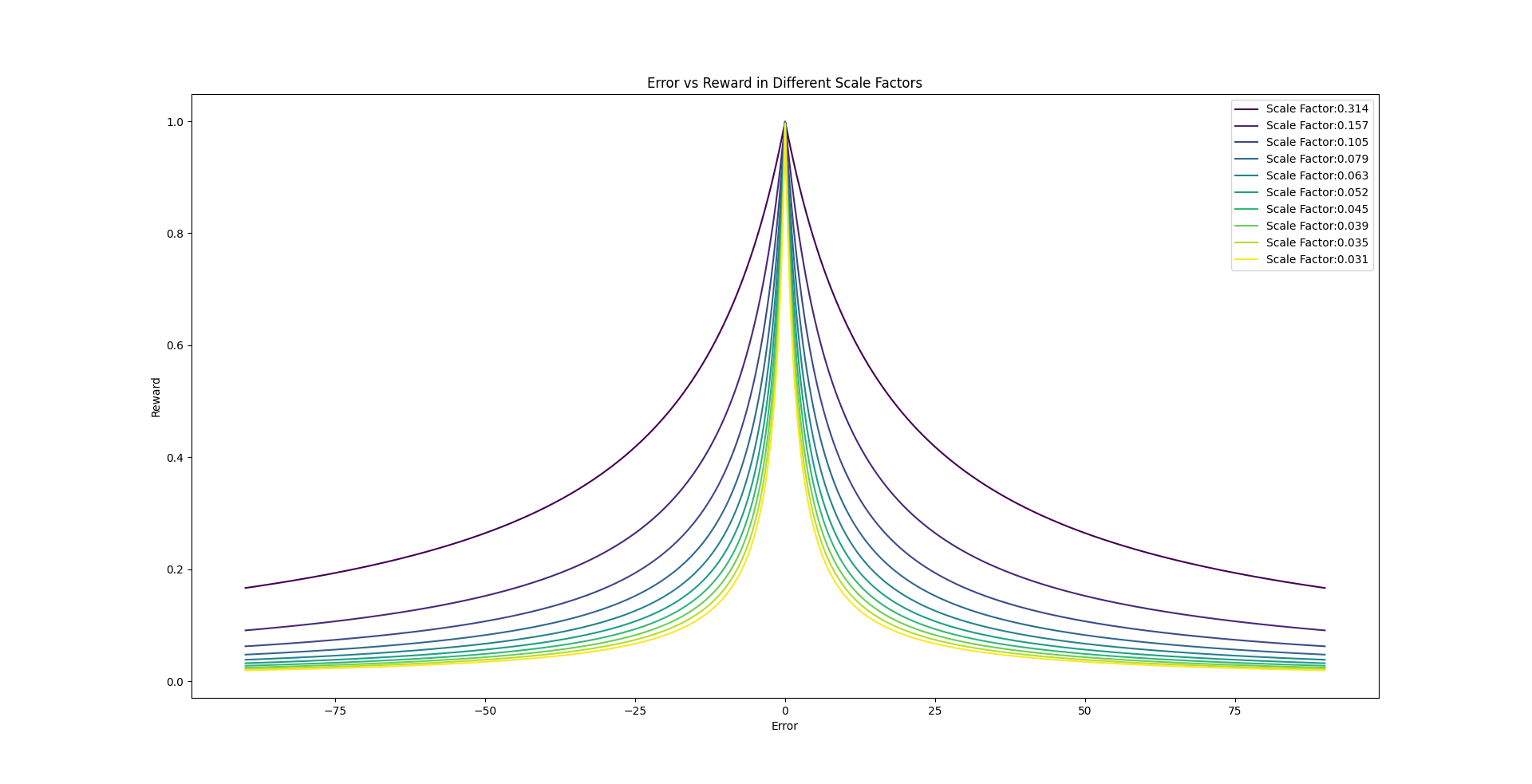}}
\caption{The different results of reward w.r.t scale factor, based on roll difference for asymptotic error component formulations.}
\label{scale}
\end{figure}

\[
e_{\text{asymp}} = \frac{e_\text{scaled}}{1 + e_\text{scaled}}
\]

The error is normalized between 0 and 1 in the last operation. And the respective error calculation is finalized. Afterwards, the reward for the specific error is calculated with substracting the error from 1. Total reward is calculated with weighted average, where the weights are selected with heuristic calculations.

For the action errors, the linear error component is used since asymptotic aspect is harmful to it. The acceleration of the error should be constant whether the error value is big or small to have smooth learning. 

The linear error component has similar formulation like asymptotic one, the difference being instead of scale factor and normalization steps, two steps are combined as scale factor being maximum error, determibed by aircraft's properties, effectively normalizing it.

In Figure \ref{model}, network architecture that is used to achieve this objective is depicted. This network is called policy network, in which we also refer it as learned controller. In optimization of this network w.r.t our objective, we have used different RL algorithms such as PPO, SAC and TD3. We have seen that the models trained with SAC algorithm outperformed the other ones, thus we have chosen SAC as our main algorithm. We have picked state variables and action variables according to the previous works of PARS literature. It is important that our learned controller needs to have same inputs and outputs with the classic controller. By ensuring that, we can make fair comparisons between two models. 

\begin{figure}[htbp]
\centerline{\includegraphics[width=0.52\textwidth]{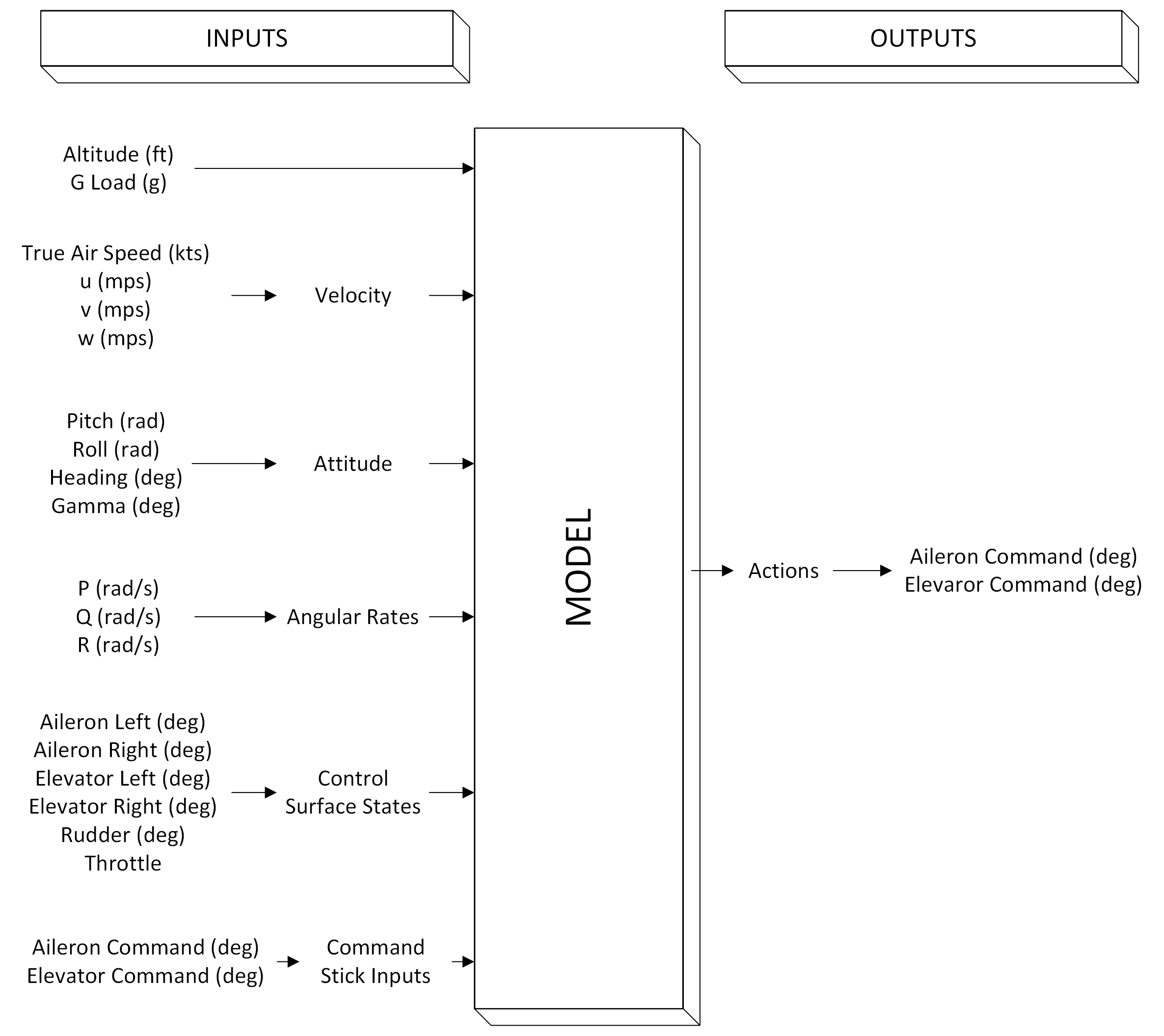}}
\caption{The general model scheme for PARS. Model is given as a black-box. The inputs are shown in the left side of the model, with different categorizations. The outputs, which are the stick commands are on the right.}
\label{model}
\end{figure}

Figure \ref{model} shows the state space and action space for our RL model. The state space is cultivated with considering necessary aircraft conditions, that is, the RL model should be able to execute the PARS task with knowing these states only. We have also minimized the state space to add less complexity into model, which also impacts training time. 
Our action space consists of only aileron and elevator stick commands. Though rudder stick command and throttle are  also available to add as stick commands, the domain experts commented that the PARS task should be executable with former two commands only, and we designed our action space with that in mind.

Our RL pipeline utilizes two important techniques for RL and hyper-parameter optimisation, namely off-policy algorithms and Optuna. 

First, as stated previously, the black-box RL model shown in Figure \ref{fig} is trained via off-policy SAC algorithm \cite{sac-paper}. Noticing that we are trying to model a controller for a specific task, on-policy algorithms first find a potential controller, then try to improve it. That may potentially cause the controller fell into local optima. On the other hand, off-policy algorithms try to cover whole space via not using a potential solution as base, which is better regarding our problem.

Secondly, hyper-parameters of an AI algorithm plays crucial part. To find the hyper-parameters that gives best results, many trials and errors are necessary. Some algorithms such as grid search were developed to ease this trials. One of the latest method for this task is called Optuna \cite{akiba2019optuna}, where the previous trials are used to determine next hyper-parameter, which searches the hyper-parameter space more efficiently.

For finding  the hyperparamaters of the SAC, using the Optuna as search method pipeline, below table depicts the determined hyperparameters:

\begin{table}[htbp]
\caption{Hyperparamaters}
\begin{center}
\begin{tabular}{|c|c|}
\hline
\textbf{Name} & \textbf{Values} \\
\hline
Batch Size & 128 \\
\hline
Buffer Size & 10000 \\
\hline
Gamma & 0.9 \\
\hline
Learning Rate & 0.000483785052402479 \\
\hline
Learning Starts & 10000 \\
\hline
Log Std Init & -2.193334342451813 \\
\hline
Net Arch & [64,64] \\
\hline
Tau & 0.08 \\
\hline
Train Freq & 512 \\
\hline
\end{tabular}
\label{tab1}
\end{center}
\end{table}

To test and visualize the model, we have chosen FlightGear as our simulator. RL environment is validated via using modified gym JSBSim-F16 FDM \cite{gym-jsbsim} and actual training is done with Hürjet FDM provided in .dll format. These topics are explained in detail in results section.

\subsection{Iterative Reward Shaping}
Below, four sections shows the iterative process of reward function design.

\subsubsection{Base Errors}\label{AA}
As the simplest attempt, we defined the terminal condition for PARS and constructed a reward function with these base components. Since the aim of PARS is recovering to level flight, we can accept the terminal condition as;

\begin{displaymath}
\phi =0 , \gamma =0
\end{displaymath}

Considering that, we defined the reward for the agent as subtracting the weighted error of $\phi$ and $\gamma$ angle from 1, with using asymptotic reward function scheme. It should be noted that the initial conditions are selected so that the whole possible upset state space is covered. 

\subsubsection{}{Base Errors with commands}\label{AA}
We have observed that, with only using base errors, actions are oscillated after some point. To solve this undesired behaviour, we added punishments to the action derivatives. Positive rewards encourage the agent to react to the level flight condition. Negative rewards(punishments) decreases the reward in high command derivative cases, thus dictates the agent to use smoother actions.

Now, the reward system is divided as positive and negative rewards as follows

Positive rewards:
\begin{displaymath}
\phi =0 , \gamma =0
\end{displaymath}

Negative rewards:
\begin{itemize}
  \item Aileron command derivative: 0 for lowest punishment
  \item Elevator command derivative: 0 for lowest punishment
\end{itemize}

To calculate total reward, each reward and punishment component is multiplied by a weight factor. Increasing the wight factor of command derivatives results with smoother action, but increasing them too much causes the agent to not act at all. Therefore, a proper weight factor should be determined to satisfy optimum point of aforementioned trade-off. Moreover, to dentote equal punishment for oscillations, we have used linear reward scheme for derivatives.

\subsubsection{Sequential errors}\label{AA}

In previous implementations, $\phi$ and $\gamma$ errors have the same weight factor. That implies that they are equally significant for the agent. To add more freedom and complexity, we have changed our approach to use sequential errors. The sequential errors are defined such that a error component may be dependent on another error component. Therefore, it is possible to describe various behaviours as a reward function. In our case, there are two possibilities:

\begin{itemize}
  \item Initially recover $\gamma$ angle, then recover $\phi$ angle
  \item Initially recover $\phi$ angle, then recover $\gamma$ angle
\end{itemize}

The max values for rewards and punishments are exactly the same as previous schemes. The procedure for reward calculation is also similar, except for sequential rewards. For sequential components, main reward is multiplied with the dependent components and a weight actor. For the first case, $\gamma$ is dependent on the $\phi$ and vice versa for second case.

During training, this configuration did not converged to any acceptable policy (model), even with hyper-parameter optimization. Trained models are not able to recover $\gamma$ angle without recovering $\phi$ angle.Therefore, we concluded that utilizing sequential errors/rewards does not provide significant improvement on PARS.

\subsubsection{Negative g Limitations }\label{AA}

After we have examined the previous implementations, we have seen that pilot is frequently exposed in negative g force. Considering that being exposed to g forces below -2g can be lethal, we introduced g limitations to the reward function. In short, when agent is encountered with g exposition lower than -2g, the episode is terminated.

The rewards are similar to the previous implementations, with $\phi$ error is defined as sequential error dependent to the $\gamma$ angle, i.e. the agent tries to recover $\phi$ angle after it recovered the $\gamma$ angle.

We have inspected two cases for this. In Case 1, the plane had started with nose up condition. previous generation agent was recovering $\phi$ and $\gamma$ angles at the same time, hence resulting in negative g exposure. Adding the negative g condition, recent model learned to recover $\gamma$ angle by maintaining $\phi$ angle between 90 and 270 degrees, then recovered the $\phi$ angle to level flight. As a result, agent experienced minimal negative g exposure which can be seen in the figures in result section.

In Case 2, nose down condition, recovery maneuver is similar to previous model since negative g exposure is not useful or necessary for recovering from similar initial conditions.

\section{Results}

As we have explained in the methodology section, we have determined that sequential reward function does not improve the performance of more than the second iteration. So, base errors with command derivatives and g errors are the best case to use. Below table denotes the scaling factors and weights for each error unit:

\begin{table}[htbp]
\caption{Error Weights and Scales}
\begin{center}
\begin{tabular}{|c|c|}
\hline
\textbf{Name} & \textbf{Values} \\
\hline
$\phi$ scale factor & 0.157 \\
\hline
$\phi$ weight & 0.25 \\
\hline
$\gamma$ scale factor & 4.5 \\
\hline
$\gamma$ weight & 0.75 \\
\hline
Aileron command derivative weight & -0.1 \\
\hline
Elevator Command Derivative Weight & -0.1 \\
\hline
\end{tabular}
\label{tab1}
\end{center}
\end{table}

After determining that, we have compared the RL based PARS and classical controller based PARS with the domain experts. Both classical control systems and RL based PARS is evaluated for whole state space, and domain experts determined that RL based PARS give better timing response for the recovery maneuvers, as it is the most important metric to compare with. Below graphics will provide the comparison metrics for two cases. 

Figure \ref{case1fig} shows the recovery maneuvers for the case where the initial values are:

\begin{displaymath}
\phi =-100 , \gamma =45
\end{displaymath}

From the graphs, we can denote two important differences. First, we can see that the AI maneuver recovered both $\phi$ and $\gamma$ faster. The classical controller recovered the $\phi$ in around 8 seconds, whereas the AI model recovered in around 6 seconds. The disparity in $\gamma$ is more extreme, where the classical controller was not able to fully recover $\gamma$ to 0 degrees, but the AI model managed to do it in around 8 seconds.

The second important observation is the difference between the g forces that the pilots take. During the recovery of the classical controller, the negative g force affects the pilots, whereas in the AI case, this is not observed.

Both of these remarks can be explained via the recovery maneuver method AI had taken. Due to the reward/punishment scheme, while recovering the $\gamma$, the AI first inversely flies the plane for avoiding negative g force, then recovers the $\phi$ angle when the desired $\gamma$ angle is achieved. The video provided on \cite{youtube2021} shows this more clearly.

\begin{figure}[htbp]
\centerline{\includegraphics[width=0.50\textwidth]{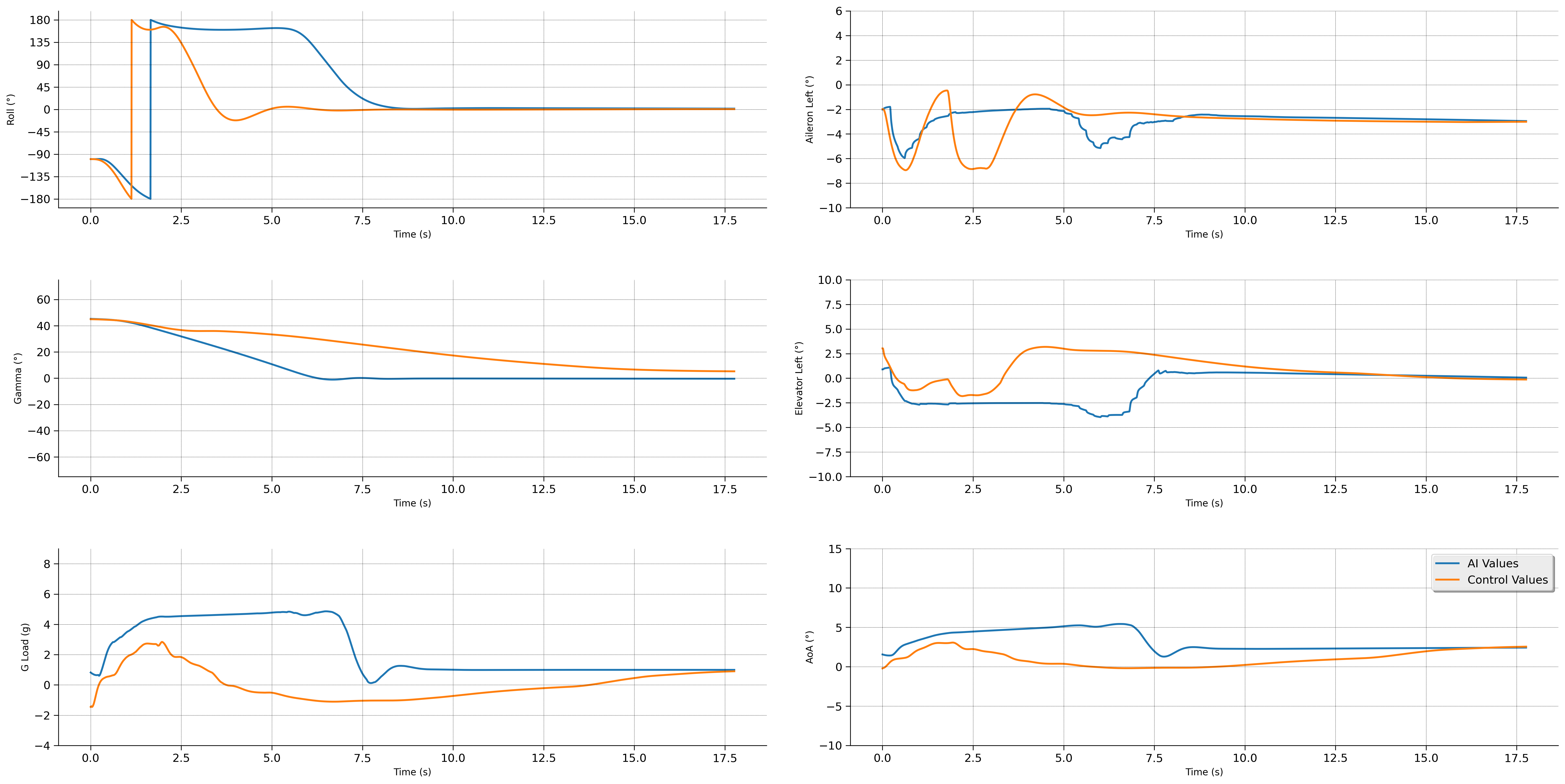}}
\caption{The PARS model comparison for case 1}
\label{case1fig}
\end{figure}

Figure \ref{case2fig} shows a similar maneuver, with different initial conditions. In this case, initial values are:

\begin{displaymath}
\phi =-30 , \gamma =60
\end{displaymath}

The roll recovery is obtained around 8 and 12 seconds, in AI and classical controller, respectively. The full $\gamma$ recovery still did not happen in classical controller, whereas AI model managed it in 10 seconds. The -g situation still exists in this case, too.

Due to $\gamma$ angle being higher in this case, the inverted flight duration is much faster, which can be observed in the graph and shown more in the video.

\begin{figure}[htbp]
\centerline{\includegraphics[width=0.50\textwidth]{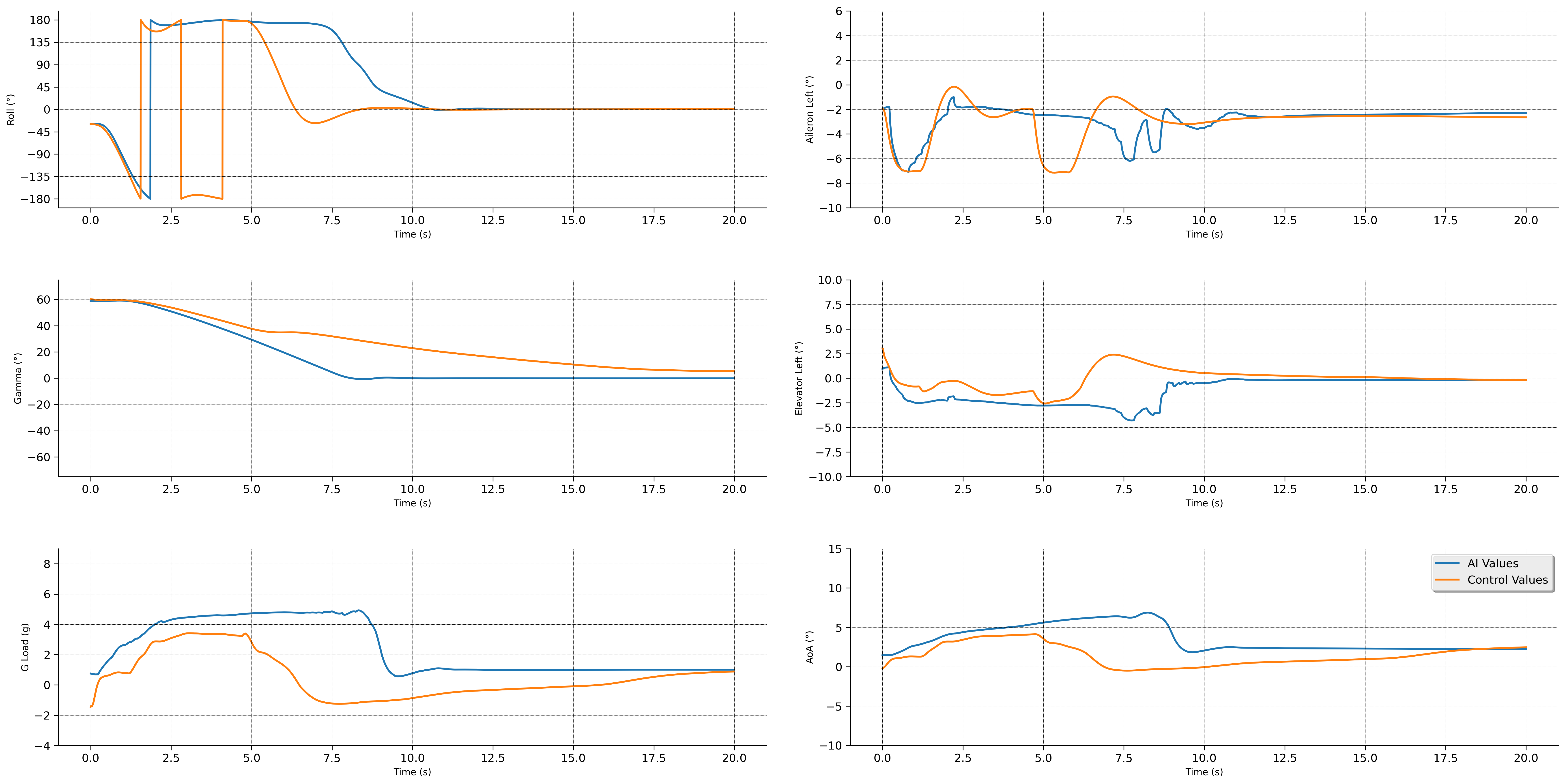}}
\caption{The PARS model comparison for case 2}
\label{case2fig}
\end{figure}

For a better visualization of the results, please check the videos at \cite{youtube2021}

\section{Conclusion}
This paper uses RL to optimize an unconstrained PARS controller. The work of incorparating avionic domain knowledge into the state-of-the-art RL models were realized.

The state-of-the-art RL models with the incorporation of avionic domain knowledge were worked with. Different reward schemes were developed to optimize the problem.
For the training, an advanced jet trainer model was used in the simulation environment. The RL model was trained freely without any external constraints with the resulting model was evaluated and compared by the domain experts. The domain experts denoted that the RL model generates optimal recovery maneuvers, outperforming classical control methods. Also, a free RL model provided the ability to cover the whole upset condition space than previous methods.

\section*{Acknowledgement}
The project was realized as part of Hurjet advanced jet trainer platform's artificial intelligence concept projects within Turkish Aerospace. The domain experts were employees of the company. The simulation environment and flight dynamics model of the jet were provided from the company.

\bibliographystyle{ieeetr}
\bibliography{bibtex}

\vspace{12pt}

\end{document}